\documentclass[letterpaper, 10 pt, conference]{ieeeconf}
\pdfoutput=1
\usepackage{times}
\usepackage{graphicx}
\usepackage{amsmath,amssymb,amsopn,amstext,amsfonts}
\usepackage{cancel}
\usepackage[space]{cite}
\usepackage{pdfsync}
\usepackage{balance}
\usepackage{color}
\usepackage{mathtools}
\usepackage{algpseudocode}
\usepackage{algorithm} 
\usepackage{bm}

\usepackage{diagbox}
\usepackage{float}
\usepackage{epstopdf}
\usepackage{pifont}
\usepackage{fixltx2e}
\usepackage{amsmath}
\usepackage{multirow}
\usepackage{url}
\usepackage{verbatim}
\makeatletter
\let\NAT@parse\undefined
\makeatother
\usepackage[linkcolor=black,citecolor=black,urlcolor=black,colorlinks=TRUE]{hyperref}
\usepackage{booktabs}
\usepackage{graphicx}
\usepackage{subcaption}
\usepackage{gensymb}
\graphicspath{{./figures_diffusion/}}
\DeclareGraphicsExtensions{.png,.jpg,.eps,.pdf}

\IEEEoverridecommandlockouts
\begin{document}

	\title{\LARGE \bf Towards Dense and Accurate Radar Perception Via \\ Efficient  Cross-Modal   Diffusion Model}
	\author{Ruibin Zhang$^{*1, 2}$, Donglai Xue$^{*2}$, Yuhan Wang$^{3}$, Ruixu Geng$^{4}$, and Fei Gao$^{\dagger{1, 2}}$
		\thanks{ $^*$Indicates equal contribution. }
		\thanks{$^\dagger$Corresponding author: {\tt\small fgaoaa@zju.edu.cn}.}
		\thanks{
			$^1$Institute of Cyber-Systems and Control, College of Control Science and Engineering, Zhejiang University, Hangzhou 310027, China.} 
		\thanks{$^2$Huzhou Institute, Zhejiang University, Huzhou 313000, China.
		}
		\thanks{$^3$S-Lab, Nanyang Technological University, Singapore 639798.
		}
		\thanks{$^4$School of Cyber Science and Technology, University of Science and Technology of China, Hefei 230026, China.
		}
	}

	\maketitle
	\thispagestyle{empty}
	\pagestyle{empty}

	\begin{abstract}
		\label{sec:abstract}\textbf{
			Millimeter wave (mmWave) radars have attracted significant attention from both academia and industry due to their capability to operate in extreme weather conditions. However, they face challenges in terms of sparsity and noise interference, which hinder their application in the field of micro aerial vehicle (MAV) autonomous navigation. To this end, this paper proposes a novel approach to dense and accurate mmWave radar point cloud construction via cross-modal learning. Specifically, we introduce diffusion models, which possess state-of-the-art performance in generative modeling, to predict LiDAR-like point clouds from paired raw radar data. We also incorporate the most recent diffusion model inference accelerating techniques to ensure that the proposed method can be implemented on MAVs with limited computing resources. We validate
			the proposed method through extensive benchmark comparisons and real-world experiments, demonstrating its superior performance and generalization ability. 
			Code and pre-trained models will be available at   https://github.com/ZJU-FAST-Lab/Radar-Diffusion.
		}
		
	\end{abstract}

	\IEEEpeerreviewmaketitle

	\section{Introduction}
	\label{sec:introduction}	
	
	In recent years, mmWave radars have shown promising results in various areas such as security, health care, and robotics. Especially in advanced driver assistance systems (ADAS), mmWave radars have emerged as an auxiliary sensing modality, serving as an alternative to optical range sensors such as LiDARs and stereo cameras in visually degraded environments.
	In contrary to LiDARs and cameras, mmWave radars cannot directly provide dense and accurate environmental information since they inherently suffer from poor angular resolution and sensor noise. Conventional mmWave radar signal processing algorithms, such as constant false alarm rate (CFAR) \cite{richards2005fundamentals}, can only generate sparse point clouds that can hardly be used for simultaneous localization and mapping (SLAM). 
	To better utilize mmWave radars, the typical technological approach is to fuse filtered radar point clouds with other sensors such as IMUs and stereo cameras for tasks like odometry\cite{park20213d, zhang20234dradarslam, huang2023multi} and object detection\cite{chadwick2019distant, cheng2021robust, wang2021rodnet}, which do not necessarily require dense mapping. For ADAS, this is sufficient because cars often drive in open environments where knowing the presence of surrounding objects and their relative distances can ensure safe driving. However, for MAVs autonomous flying in cluttered environments, accurate scene mapping is necessary, which is tough for mmWave radars. Furthermore, due to payload limits, most MAVs can only carry single-chip mmWave radars with angular resolutions approximately $\mathbf{1\%}$ of that of LiDAR, which poses even greater challenges.  
	
	To obtain richer environmental information via mmWave radars, several recent studies propose to generate pseudo-LiDAR point clouds from raw radar data via deep learning methods \cite{lu2020see, cheng2022novel, prabhakara2023high, geng2023dream}.
	These results are achieved by cross-modal supervision learning in which LiDAR point clouds are typically adopted as the ground truth.
	However, the enhanced radar point clouds obtained from these methods are still inferior to LiDAR point clouds in both density and accuracy, and cannot meet the requirements of MAV autonomous navigation.
	
	\begin{figure}
		\centering
		\includegraphics[width=1\linewidth]{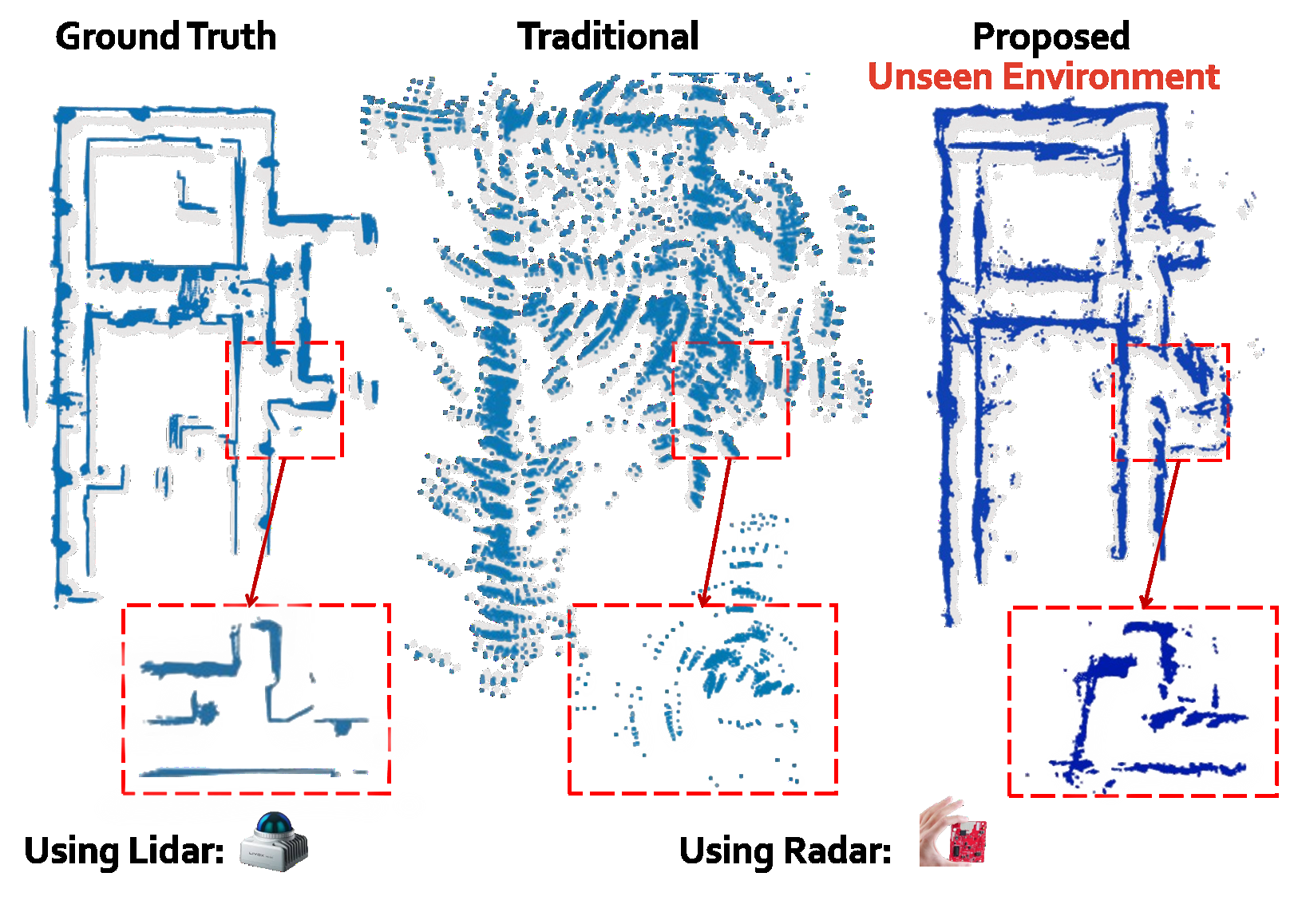}
		\captionsetup{font={small}}
		\caption{
			A reconstruction quality comparison carried out at Huzhou Institute of Zhejiang University. Compared to the dense and accurate mapping results of LiDAR, a single-chip mmWave wave radar can only generate sparse and noisy point clouds through traditional radar target detectors. However, with the proposed diffusion model-based method, we are able to generate radar point clouds that are close to the ground truth in \textbf{one-step generation}. Moreover, the model we use is trained on a public dataset with \textbf{completely different} scenes and sensor configurations, revealing the generalization ability and robustness of our method. 
		}
		\label{pic:realworld_test}
		\vspace{-1.5cm}
	\end{figure}

	\begin{figure*}[t]
		\centering
		\includegraphics[width=1\linewidth]{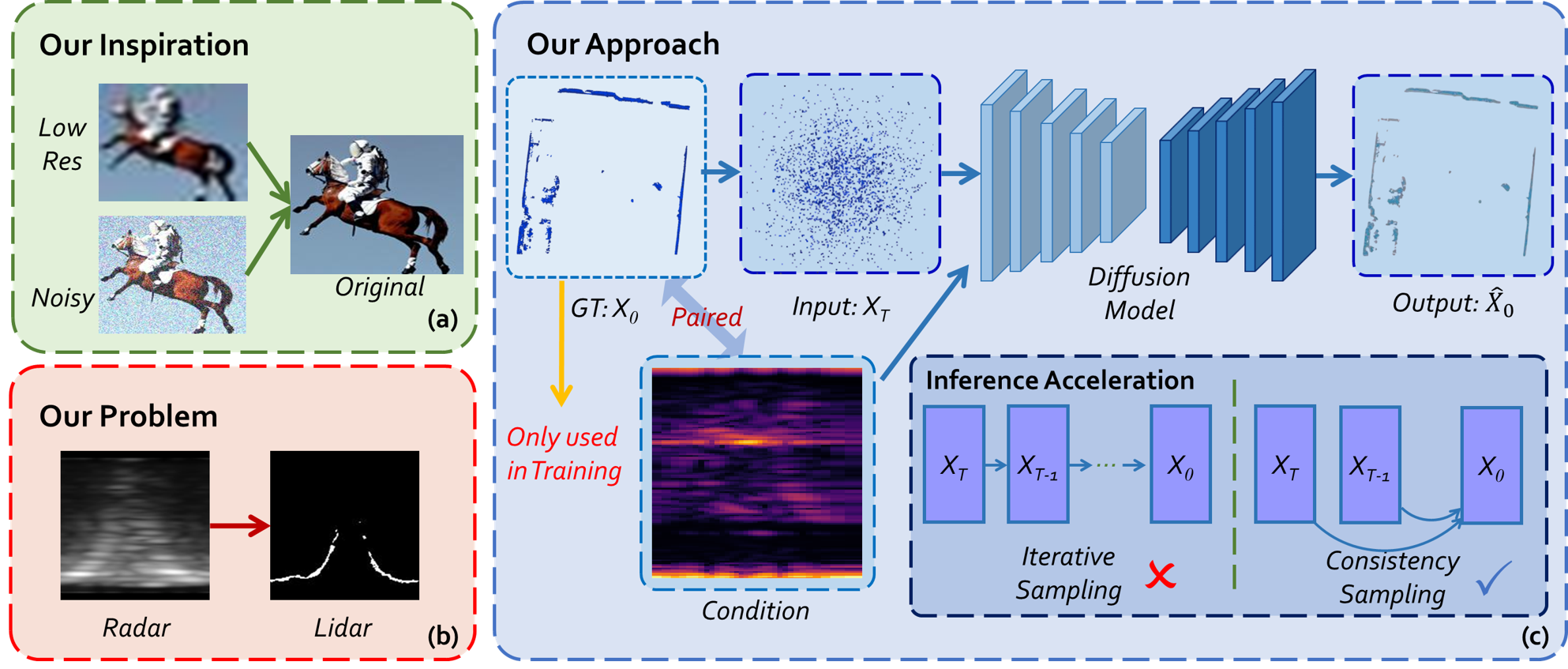}
		\captionsetup{font={small}}.
		\caption{
			(a)  the image restoration tasks where low-resolution and noise-corrupted images can be restored to ground truth images using neural networks. (b) Spatially and temporally aligned radar range-azimuth heatmap (RAH) and LiDAR bird's eye view (BEV) image expressed in polar coordinate. As can be seen, the angular resolution of millimeter-wave radar is considerably lower than that of LiDAR, and it is heavily affected by noise. Therefore, predicting LiDAR BEV images from paired RAHs can be modeled as image restoration. (c) Diagram of the architecture of our proposed approach. During training, ground truth Lidar point clouds $\mathbf{x}_0$ are corrupted to $\mathbf{x}_t$ through diffusion process, then the neural network is trained to estimate the ground truth conditioned on the paired radar RAHs. During inference, The neural network directly predicts $\hat{\mathbf{x}_0}$ from pure Gaussian noise $\mathbf{x}_T$ and radar RAHs. To resolve the iterative sampling issue in diffusion models, we incorporate consistency models\cite{song2023consistency} that enable one-step generation from $\mathbf{x}_T$ to $\hat{\mathbf{x}_0}$ in our approach.
		}
		\label{pic:inspiration}
		\vspace{-0.7cm}
	\end{figure*}

	
	In this work, we aim to achieve dense and accurate environmental perception mmWave radar for MAV autonomous navigation in visually challenging environments. We present a novel learning-based approach for single-chip mmWave radar point cloud generation that achieves state-of-the-art performance with point cloud density and accuracy comparable to that of LiDAR. Since mmWave radar signal is sparse and noisy, generating high-quality radar point clouds through LiDAR point cloud supervision is essentially a cross-modal denoising and super-resolution task, which places high demands on the generative capability of deep learning methods.  
	To resolve this, we adopt diffusion models, which have emerged as the new state-of-the-art family of deep generative models by virtue of their strong expressiveness and training stability over generative adversarial networks (GANs) \cite{goodfellow2020generative} and variational autoencoders (VAEs) \cite{kingma2013auto}.
	When applied to the field of robotics, diffusion models can effectively learn and represent complicated relationships between different modalities of data, such as environmental features and robot dynamics\cite{carvalho2023motion, janner2022planning}. 
	To incorporate diffusion model in our task, we formulate the original problem as image restoration, which learns the mapping between the "impaired" (radar) and the "original" (LiDAR) image domains.
	A limitation of diffusion model is its dependence on an iterative sampling process that causes slow inference. To achieve real-time performance when applied to MAVs with constrained computing capacity, we combine the diffusion model inference accelerating technology that supports one-step generation in our method.
	
	To verify the proposed method, we conduct extensive benchmark comparisons against both conventional and recent learning methods on a public dataset including various real-world environments. The results demonstrate that our method exceeds the previous works in radar point cloud quality. Moreover, we carry out tests with our self-made dataset to demonstrate that the proposed method can generalize to cases with unseen environments and diverse sensor configurations. The contributions of this paper are as follows:
	
	\begin{itemize}
		\item [1)] 
		A novel mmWave radar perception enhancing method that can generate dense and accurate LiDAR-like point clouds. To the best of our knowledge, we are the first to introduce diffusion models into mmWave radar point cloud generation for MAV autonomous navigation.
		\item [2)]
		Incorporating diffusion model inference accelerating methods to achieve real-time performance on embedded computing platforms that can be mounted on MAVs.
		\item [3)]
		A set of real-world tests and benchmark comparisons
		that validate the proposed method. Moreover, we will open source the codes and pre-trained models for the reference of the community. 
	\end{itemize}	
	
	\section{Related Work} 
	\label{sec:related_works}	
	MmWave radars transmit millimeter-wavelength electromagnetic waves and capture the reflected signals that can determine the range, velocity, and angle (in both azimuth and elevation direction) of the surrounding objects. Millimeter waves can penetrate fine particles such as raindrops, snowflakes, and dust, allowing mmWave radars to operate under various harsh weather conditions. However, they are significantly impacted by multipath effects and clutter interference, resulting in considerable noise in the reflected signals. Only sparse and noisy radar point clouds can be obtained through conventional target detectors such as CFAR\cite{richards2005fundamentals} and MUSIC\cite{schmidt1986multiple}.
	
	As stated in Section \ref{sec:introduction}, some researchers focus on radar point cloud post-processing for tasks such as odometry and object detection. Park et al. \cite{park20213d} first use random sample consensus (RANSAC) to filter outlier radar point clouds and then achieve ego-velocity estimation utilizing the relative velocity estimation from radars. Afterwards, they adopt pose-graph SLAM with this velocity factor to estimate ego motion. Some subsequent works\cite{zhang20234dradarslam, huang2023multi} also adopt this technical scheme. Chadwick el al.\cite{chadwick2019distant} first propose to fuse radar and monocular images for object detection using a convolutional neural network. Cheng et al. \cite{cheng2021robust} propose a radar-image spatiotemporal fusion network to
	fuse adjacent frames radar data with single frame RGB image and achieve small object detection on the water surface. Ding et al.\cite{ding2023hidden} propose a scene flow estimation method based on radar point clouds using cross-modal supervision from co-located heterogeneous sensors (odometer, LiDAR, and camera). All the above works are based on sparse point clouds generated by conventional radar signal processing methods, which cannot achieve dense mapping. 
	
	Lu et al.\cite{lu2020see} are the first to propose a learning-based radar point clouds upsampler through cross-modal supervision from a co-located LiDAR. This method does not utilize the raw data from the mmWave radars, instead using the sparse point clouds generated by conventional radar signal processors, in which much information is lost. Therefore, the effectiveness of this method is limited. The method proposed by Mopidevi et al.\cite{mopidevi2023rmap} faces the same issue as well. Cheng et al.\cite{cheng2022novel} use LiDAR point clouds as supervision to filter out noise in the radar range-doppler heatmaps (RDHs) generated  with a low threshold, and then use conventional direction of arrival (DOA) estimation methods to obtain point clouds. As this method only uses LiDAR as a filter, the perception characteristics of LiDAR are not learned. In addition, this method still resorts to conventional DOA methods for radar point cloud generation. As a result, the point clouds obtained by this method are not dense and accurate enough to depict the surrounding environment. Prabhakara et al.\cite{prabhakara2023high} attempt to learn the mapping from radar RAHs to LiDAR point clouds through a U-net based architecture. However, due to the limited generative capability of the network architecture, the resulting radar point clouds lose sharp edges and detailed texture features. Geng et al. \cite{geng2023dream} first adopt non-coherent accumulation and synthetic aperture accumulation to improve the density and angular resolution of radar point clouds, then use neural networks to filter the noise and interference under the supervision of LiDAR point clouds. Nonetheless, the non-coherent accumulation and synthetic aperture accumulation methods require accurate ego-motion estimation which is not available in extreme environments where other sensor modalities fail. 
	
	In summary, the aforementioned methods either cannot accurately perceive the environment or rely on state estimation provided by other sensors, making them inapplicable to MAV autonomous navigation in harsh environments.

	
	\section{Preliminaries}
	\label{sec:Preliminaries}
	\subsection{MmWave Radar Signal Processing}
\begin{figure}
		\centering
		\includegraphics[width=0.85\linewidth]{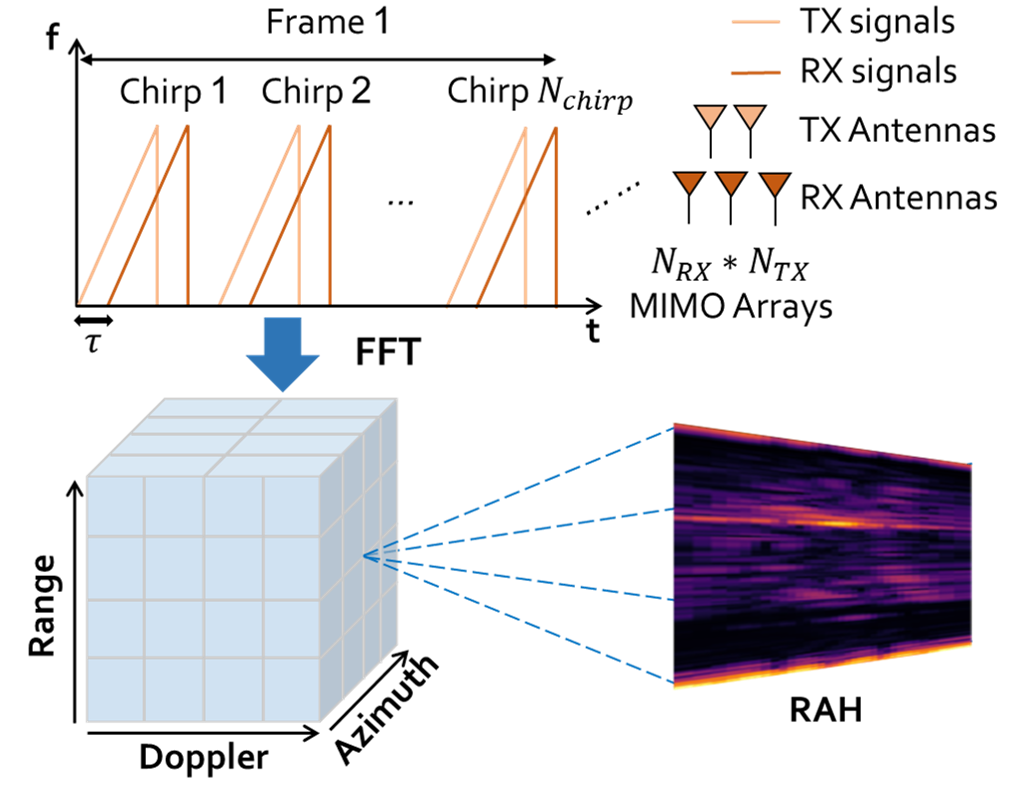}
		\captionsetup{font={small}}
		\caption{
			Illustration of the mmWave radar signal model and data preprocessing. In the radar data cube, the elevation dimensions is omitted for the convenience of visualization.
		}
		\label{pic:radar_principle}
		\vspace{-1.2cm}
	\end{figure}

	Modern commercial mmWave radars mostly transmit frequency-modulated continuous wave (FMCW) chirps, in which the frequency linearly increases over time. A complete radar frame consists of $N_{chirp}$ chirps with equal time intervals. With the help of multiple-input and multiple-out (MIMO) technology, a radar with $N_{TX}$ transmit (TX) antennas and $N_{RX}$ receive (RX) antennas can produce at most $N_{TX}*N_{RX}$ virtual antenna arrays. Range, velocity, and angle estimation can then be achieved through the following methods, as illustrated in Fig.\ref{pic:radar_principle}.
	\subsubsection{\textbf{Range Estimation}} After RX antennas capture the reflected signals from surrounding objects, a signal mixer combines the TX and RX signals to synthesize an intermediate frequency (IF) signal whose frequency is equal to the frequency difference of the TX and RX signals. For FMCWs whose frequency linearly increases with time, the frequency of an IF signal is constant and is equal to the inverse of the time difference between the TX and RX signals (denoted as $\tau$). Therefore, we can calculate the range of detected objects as $r=\frac{C\tau}{2}=\frac{Cf}{2S}$, where $C$ is the speed of light, $S$ is the FMCW slope rate, and $f$ is the frequency of IF signal determined by fast Fourier transform (FFT).
	
	\subsubsection{\textbf{Velocity Estimation}} The relative velocity between the radar sensor and the surrounding objects is measured via the phase difference between different chirps: $v=\frac{\lambda \Delta \Phi_v}{4 \pi T_c}$, where $T_c$ is the time duration of a chirp, $\Delta \Phi_v$ is the phase difference determined by a second FFT called Doppler FFT. 
	
	\subsubsection{\textbf{Angle Estimation}} The estimation of the azimuth and elevation angle of the surrounding objects are based on the fact that the different ranges from the objects to different RX antennas bring about a detectable phase difference. Both the azimuth and elevation angle of arrival can be calculated by: $\theta=\sin ^{-1}\left(\frac{\lambda \Delta \Phi_a}{2 \pi l}\right)$, where 
	$\lambda$ is the wavelength, $l$ is the distance between the RX antennas ($l$ equals to $\frac{\lambda}{2}$ usually), $\Delta \Phi_a$ is the phase difference determined by a third FFT called Angle FFT. It is worth mentioning that the angular resolution is $\theta_{res}=\frac{\lambda}{N_{TX}*N_{RX} \cdot l \cdot \cos \theta}$. For a single chip radar with $3$ TX and $4$ RX antennas ($8$ virtual arrays in azimuth and $2$ in elevation), it has angular resolutions of only $14^\circ$ and $57^\circ$ in the azimuth and elevation directions, respectively.
	
	\subsubsection{\textbf{Point Cloud Generation}} After applying multiple FFTs along range, velocity, and angle (both azimuth and elevation) dimension, the raw data is transformed to a 4D tensor in which any two dimensions can be combined to create a continuous heatmap (e.g., RDH and RAH). Commonly, conventional target detectors such as various kinds of CFAR are then applied to extract valid targets against noise and interference. CFAR detectors dynamically estimate the noise level based on the surrounding cells of the target cell, and classify cells with signal intensity higher than the noise level as valid targets. Due to the diverse sources of noise and interference in cluttered environments, CFAR cannot accurately estimate the noise level. When the threshold factor is set too low, a large amount of noise is falsely reported as targets. Conversely, setting the threshold coefficient too high results in missing many valid targets.
	\subsection{Diffusion Models}
	\label{sec:diffusion}
	Diffusion models, inspired by non-equilibrium thermodynamics, characterize the generation of data samples as a process of continuously denoising pure Gaussian noises\cite{sohl2015deep}. During training, data samples are gradually corrupted with noise according to a predefined schedule until they become indistinguishable from pure Gaussian noise. The neural network then learns the inverse process, i.e., it learns the data distribution and how to generate new samples. There are two mainstream diffusion model architectures: the denoising probabilistic diffusion model (DDPM)\cite{ho2020denoising} and the noise conditional score network (NCSN)\cite{song2019generative}. These models can be expressed within a unified score-based generative modeling framework proposed by Song et al.\cite{song2020score}.
	They formulate the diffusion process, which is denoted as $\{\mathbf{x}(t)\}_{t=0}^T$ indexed by a continuous time variable $t \in[0, T]$, as the solution to a stochastic differential equation (SDE):
	
	\begin{equation}
		\mathrm{d} \mathbf{x}=\mathbf{f}(\mathbf{x}, t) \mathrm{d} t+g(t) \mathrm{d} \mathbf{w}, 
	\end{equation}
	where $\mathbf{w}$ is a standard Wiener process, $\mathbf{f}(\cdot, t)$ and $g(\cdot)$ are the drift and diffusion coefficient of $\mathbf{x}(t)$, respectively. $\mathbf{f}$ and $g$ specify the schedule at which the original samples $\mathbf{x}(0)$ are corrupted by noise. They are pre-designed and do not contain any learnable parameters.
	
	Diffusion process holds the property that the reverse of a diffusion process is also a diffusion process, running backwards in time and given by the reverse-time SDE\cite{song2020score}:
	
	\begin{equation}
		\mathrm{d} \mathbf{x}=\left[\mathbf{f}(\mathbf{x}, t)-g(t)^2 \nabla_{\mathbf{x}} \log p_t(\mathbf{x})\right] \mathrm{d} t+g(t) \mathrm{d} \overline{\mathbf{w}}, 
		\label{eq:reverse_sde} 
	\end{equation}
	where $\overline{\mathbf{w}}$ is also a standard Wiener process, $\nabla_{\mathbf{x}} \log p_t(\mathbf{x})$ is called the score function of each marginal distribution $\nabla_{\mathbf{x}} \log p_t(\mathbf{x})$. Once $\log p_t(\mathbf{x})$ is known for all $t$, we can follow the reverse diffusion process given by Eq. \ref{eq:reverse_sde} to propagate from a random noise $\mathbf{x}_T$ to a new sample $\hat{\mathbf{x}_0}$. In diffusion models, we train the noise prediction model $\boldsymbol{\epsilon}_\theta\left(\boldsymbol{x}_t, t\right)$ to approximate $\nabla_{\mathbf{x}} \log p_t(\mathbf{x})$. For conditional generation scenarios, diffusion models learn $\boldsymbol{\epsilon}_\theta\left(\boldsymbol{x}_t, t, c \right )$, which means that given corrupted samples $\boldsymbol{x}_t$, timestamps $t$, and the corresponding conditions $c$ (such as class labels, text prompts, images, etc.). The neural networks are optimized through the following loss:
	
	\begin{equation}
		L_{D}=\mathbb{E}_{x, t, c}\left[\left\|\nabla_{\mathbf{x}} \log p_t(\mathbf{x}) - \boldsymbol{\epsilon}_\theta\left(\boldsymbol{x}_t, t, c\right)\right\|_2^2\right].
		\label{eq:denoising_loss}
	\end{equation}
	During inference, we can generate new samples through iterative forward propagation from $t = T$ to $t = 0$.

	\section{Dense and Accurate Radar \\ Point cloud Generation}
	This section elaborates on the proposed mmWave radar point cloud generation method. It is inspired by the image restoration tasks (as illustrated in Fig.\ref{pic:inspiration}a) where diffusion models achieve state-of-the-art performance\cite{kawar2022denoising}. 
	In fact, both radar RAHs and LiDAR BEV images can generate 2D point clouds in polar coordinate, but the RAHs have lower angular resolution and are corrupted with noise (as shown in Fig.\ref{pic:inspiration}b). We then propose using a diffusion model to recover the LiDAR BEV images from paired radar RAHs. The algorithm architecture is illustrated in Fig.\ref{pic:inspiration}c.
	
	\subsection{Diffusion Model-based LiDAR BEV Image Prediction}
	
	As stated in Section \ref{sec:diffusion}, the main goal of a diffusion model is to learn the reverse of the diffusion process that gradually adds noise to data samples following a predefined noise schedule.
	Given a set of spatially and temporally aligned LiDAR BEV images ${\mathbf{x}_0}$ and radar RAHs $c$, we propagate it to ${\mathbf{x}_t}$ with Gaussian noise and timestamp $t$ uniformly sampled from $\{1, . . . , T\}$. We directly predict the original data sample ${\mathbf{x}_{\theta}}({\mathbf{x}_t}, t, c)$ conditioned on $c$ instead of learning the score function,
	which is verified to be practical by Karras et al.\cite{karras2022elucidating}.
	With this training objective, we first optimize the following mean squared error (MSE) loss to have ${\mathbf{x}_{\theta}}({\mathbf{x}_t}, t, c)$ resemble ${\mathbf{x}_0}$:
	\begin{equation}
		\mathcal{L}_{m}=\left\|{\mathbf{x}_0} -{\mathbf{x}_{\theta}}({\mathbf{x}_t}, t, c)\right\|_2^2.
		\label{eq:mse_loss}
	\end{equation}
	Eq. \ref{eq:mse_loss} actually plays a similar role to Eq. \ref{eq:denoising_loss}, since $\mathbf{x}$ can be obtained from the score function $\nabla_{\mathbf{x}} \log p_t(\mathbf{x})$ via Eq. \ref{eq:reverse_sde}. It is worth mentioning that Lin et al. \cite{lin2023diffusion} report that MSE metric can lead to perceptual mismatch between the generated samples and the original samples. For example, the accuracy of the wall structures in the LiDAR BEV images in our case is more important for autonomous navigation than that of the grass. However, since the former occupies fewer pixels than the latter, the corresponding loss function is smaller. Thus, when training diffusion models only with MSE loss, it penalizes the subtle
	pixel mismatch more than prominent structural features\cite{lin2023diffusion}.
	To resolve this, we introduce the learned perceptual image patch similarity (LPIPS) metric \cite{zhang2018unreasonable} in our task. It takes the deep features extracted from neural networks as a training loss which outperforms previous perceptual metrics such as peak signal-to-noise ratio (PSNR) and structural similarity (SSIM) as reported by Zhang et al.\cite{zhang2018unreasonable}. Specifically, we pass ${\mathbf{x}_0}$ and ${\mathbf{x}_{\theta}}({\mathbf{x}_t}, t, c)$ through a pre-trained neural network $f_p$ and minimize the $\ell_2$ distance between the resulting deep features:

	\begin{equation}
		\mathcal{L}_{p}=\left\|f_p({\mathbf{x}_0}) -f_p({\mathbf{x}_{\theta}}({\mathbf{x}_t}, t, c))\right\|_2^2.
		\label{eq:perceptual_loss}
	\end{equation}
	
	Our full training loss is then written as:
	
	\begin{equation}
		\mathcal{L}_{train}= \lambda_{m} \mathcal{L}_{m} + \lambda_{p} \mathcal{L}_{p},
		\label{eq:mse_loss}
	\end{equation}
	where $\lambda_{m} = 0.8$ and $\lambda_{p} = 0.2$ are weights for each loss function. 
	
	As mentioned earlier in Section \ref{sec:diffusion}, diffusion models follow an iterative approach during inference, where each step requires forward propagation through the network. This brings huge challenges to real-time applications.
	To address this issue, we adopt the most recent consistency models \cite{song2023consistency} that support fast one-step generation while still allowing multistep sampling to trade compute for sample quality. Consistency models learn a consistency function $\boldsymbol{f}:\left(\mathbf{x}_t, t\right) \mapsto \mathbf{x}_\epsilon$, where $\epsilon$
	is a fixed small positive number. It points arbitrary pairs of $\left(\mathbf{x}_t, t\right)$ to the original sample $\mathbf{x}_0$. i.e., $\boldsymbol{f}\left(\mathbf{x}_t, t\right)=\boldsymbol{f}\left(\mathbf{x}_{t^{\prime}}, t^{\prime}\right) = \mathbf{x}_0$ for all $t, t^{\prime} \in [\epsilon, T]$. A consistency model can be distilled from a pre-trained diffusion model or trained from scratch. We choose the former approach because it results in higher sample quality. We refer the readers to \cite{song2023consistency} for more technical details. 
	
	To summarize our approach, we first pre-train a diffusion model $D_\theta$ and then distill it to a consistency model $C_\theta$ which shares the same network architecture as $D_\theta$. During inference, single-frame radar RAHs are passed through $C_\theta$ as the condition to predict paired LiDAR BEV images. Finally, we convert the BEV images to 2D point clouds with similar density and accuracy to LiDAR point clouds. 
	
	\subsection{Implementation Details}
	The neural network we adopt for our approach is built based on the U-net architecture \cite{ronneberger2015u}. It contains five downsampling and upsampling blocks, each containing three ResNet layers. The former downsamples the images from $128 \times 128$ to $4 \times 4$ to extract features, while the latter upsamples  the images from $4 \times 4$ to $128 \times 128$ to generate new samples. Similar to Nichol el al.\cite{nichol2021improved}, we add multi-head attention layers at the $32 \times 32$, $16\times 16$, and $8 \times 8$ resolution between the ResNet layers. The diffusion timestamps $t$ are passed through sinusoidal position embedding and then fed to each  ResNet layer.
	In terms of the conditioning mechanism, we resize both the RAHs $c$ and noise-corrupted LiDAR BEV images $\mathbf{x}_t$ to $128 \times 128$, and directly concatenate them to the input of the neural networks. In addition, we
	adopt the same settings of diffusion noise and timestamp schedules as EDM proposed by Karras et al\cite{karras2022elucidating}. In the EDM setup, we need 80 steps of iteration during inference to obtain the results.
	Our model has $34$M parameters in total. We use a single RTX 4090 GPU for training. $D_\theta$ is trained for 280k steps using RAdam optimizer with a learning rate of $1\text{e}-5$, and $C_\theta$ is distilled from $D_\theta$ through 140k steps of training with the same hyperparameters.

	\section{Results}
	\label{sec:results}
	In this section, we carry out benchmark comparisons and real-world tests on a public dataset and our self-made dataset to evaluate the performance of the proposed method.
	\subsection{Benchmark Comparisons}
	we conduct benchmark comparisons against three baseline methods: a conventional radar target detector OS-CFAR\cite{richards2005fundamentals}, and recent learning-based methods RadarHD\cite{prabhakara2023high} and RPDNet\cite{cheng2022novel} on the ColoRadar dataset\cite{kramer2022coloradar}. Regarding  our method, we test the results obtained by both iterative sampling of EDM\cite{karras2022elucidating} (denoted as Proposed-EDM) and one-step generation of consistency distillation\cite{song2023consistency} (denoted as Proposed-CD).
	Note that OS-CFAR and RPDNet are able to generate 3D point clouds through classical DOA estimation. However, to facilitate comparisons, we standardize the evaluation to 2D point clouds. We will explore the use of diffusion models to generate high-quality 3D radar point clouds in the future.
	
	\begin{figure}
		\centering
		\includegraphics[width=0.8\linewidth]{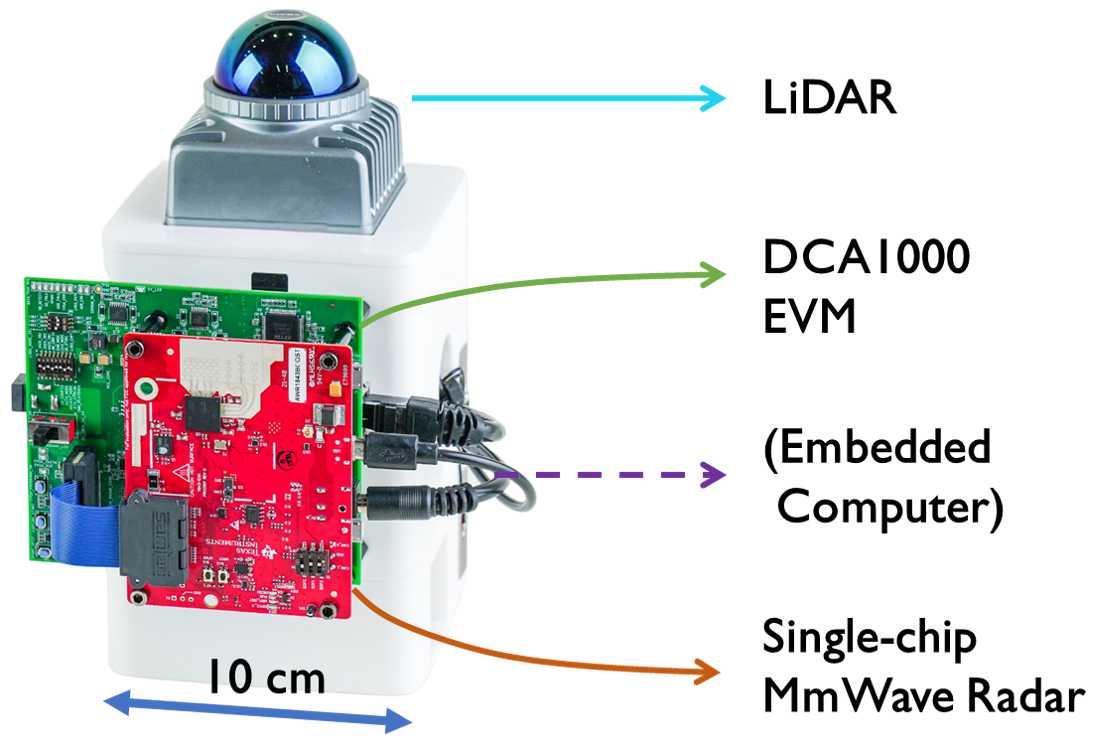}
		\captionsetup{font={small}}
		\caption{
			Our customized hand-held sensor platform. An embedded computer is placed inside it for sensor driving and data recording.
		}
		\label{pic:sensor_platform}
		\vspace{-2.8cm}
	\end{figure}
	
	\subsubsection{\textbf{Dataset Preparation}}
	\label{sec:benchmark_dataset_preparation}
	The ColoRadar dataset encompasses a total of 52 sequences of synchronized single-chip mmWave radar and LiDAR data captured in seven different scenes, including indoor (Arpg Lab, Ec Hallways and Aspen), outdoor (Longboard and Outdoors), and mine environments (Edgar Army and Edgar Classroom). We take the first three sequences of each scene as the training set and the rest as the test set (the sixth sequence of Longboard scene is dismissed because radar data is lost for 210 seconds). We first remove the LiDAR point clouds out of the field of view (FOV) of radar. Then, the annular floor and ceiling LiDAR point clouds are removed by Patchwork++ \cite{lee2022patchwork++} because they are almost unperceivable from single-chip mmWave radars.

	\subsubsection{\textbf{Qualitative Comparisons}}

	\begin{figure*}
		\centering
		\includegraphics[width=1.0\linewidth]{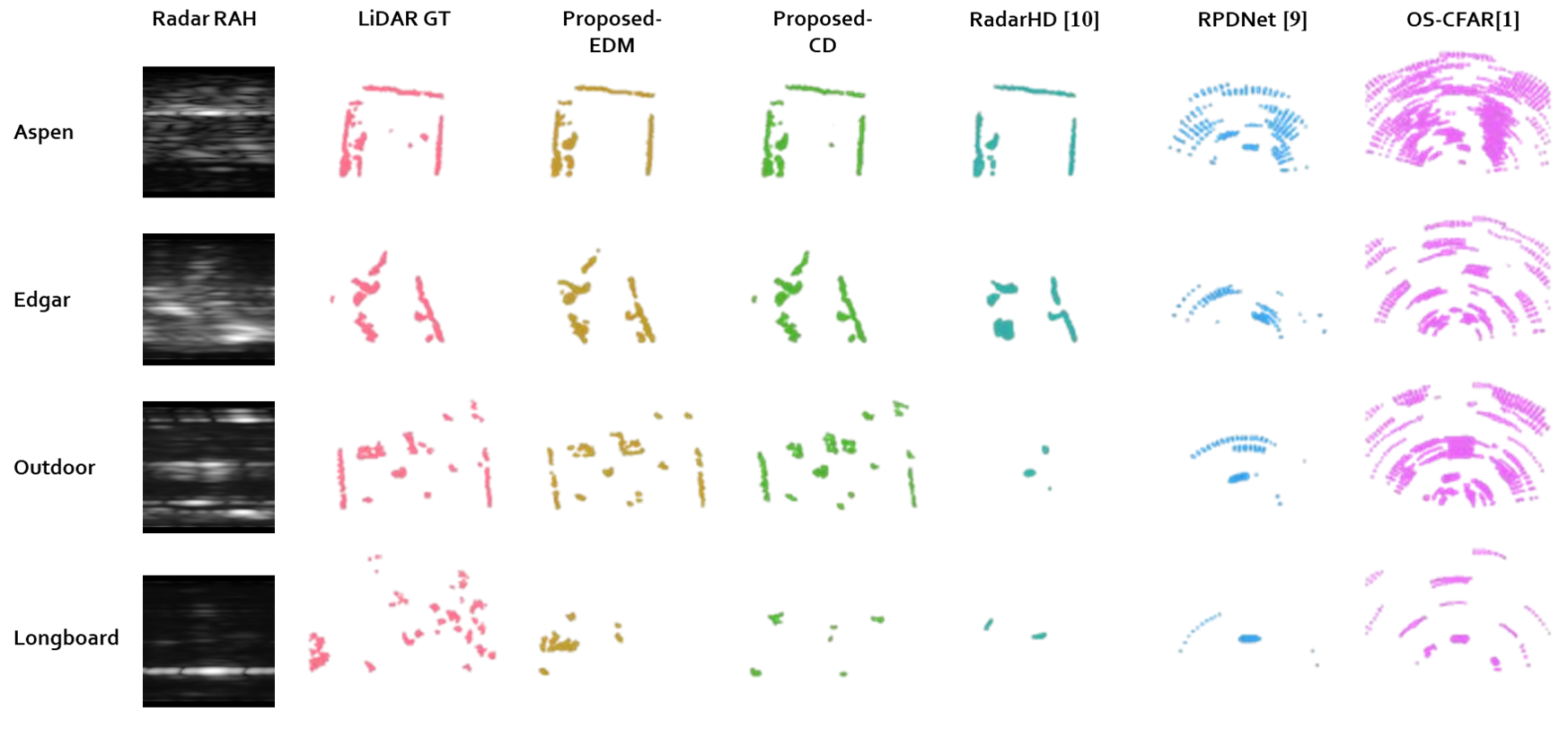}
		\captionsetup{font={small}}
		\caption{
			Qualitative comparisons of single-frame 2D point clouds conducted on the ColoRadar dataset. Typical examples of radar RAH input, LiDAR ground truth and the generated radar point clouds in the Aspen (indoor), Edgar (mine), Outdoor (outdoor), and Longboard (outdoor) scenes are shown above. The results of each method are visualized in different colors. Note that in RPDNet and OS-CFAR, the radar input also includes Doppler information. However, for the sake of visualization convenience, we only present the RAHs.
		}
		\label{pic:qualitative_results}
		\vspace{-0.1cm}
	\end{figure*}
	
	\begin{table*}[ht]
		
		\captionsetup{justification=centering} 
		\caption{Quantitative Comparisons on the Coloradar Dataset.}
		\label{tab:quantative_coloradar}
		\centering
		\renewcommand{\arraystretch}{1.6}
		\setlength\tabcolsep{10pt}
		\resizebox{\textwidth}{!}{%
			\fontsize{30}{30}\selectfont
			\begin{tabular}{l|*{6}{lll}}
				\hline
				\specialrule{0em}{6pt}{0pt}
				\multirow{2}{*}{Methods} & \multicolumn{3}{c}{Arpg Lab} & \multicolumn{3}{c}{Ec Hallways} & \multicolumn{3}{c}{Aspen}  & \multicolumn{3}{c}{Longboard}  & \multicolumn{3}{c}{Outdoors}  & \multicolumn{3}{c}{Edgar}\\
				\cline{2-19}
				& CD  \scalebox{1.2}{$\downarrow$} & HD \scalebox{1.2}{$\downarrow$} & F-Score \scalebox{1.2}{$\uparrow$} 
				& CD \scalebox{1.2}{$\downarrow$} & HD \scalebox{1.2}{$\downarrow$} & F-Score \scalebox{1.2}{$\uparrow$} 
				& CD \scalebox{1.2}{$\downarrow$} & HD \scalebox{1.2}{$\downarrow$} & F-Score \scalebox{1.2}{$\uparrow$}
				& CD \scalebox{1.2}{$\downarrow$} & HD \scalebox{1.2}{$\downarrow$} & F-Score \scalebox{1.2}{$\uparrow$}
				& CD \scalebox{1.2}{$\downarrow$} & HD \scalebox{1.2}{$\downarrow$} & F-Score \scalebox{1.2}{$\uparrow$}
				& CD \scalebox{1.2}{$\downarrow$} & HD \scalebox{1.2}{$\downarrow$} & F-Score \scalebox{1.2}{$\uparrow$}\\
				\specialrule{0em}{0pt}{6pt}
				\hline
				\specialrule{0em}{6pt}{0pt}
				OS-CFAR\cite{richards2005fundamentals} & 2.732 & 10.488 & 0.150 & 2.836 & 10.372 & 0.128 & 2.121 & 10.024 & 0.160 & 5.307 & 12.115 & 0.051 & 3.024 & 10.024 & 0.097 & 4.129 & 12.417 & 0.099 \\ 
				RadarHD\cite{prabhakara2023high} & 1.728 & 6.116 & 0.273 & 1.686 & 6.387 & 0.284 & 0.912 & 4.691 & 0.511 & 5.395 & 11.298 & $\textbf{0.149}$ & 3.096 & 10.090 & 0.245 & 0.604 & 3.070 & 0.535 \\ 
				RPDNet\cite{cheng2022novel} & 1.555 & 5.216 & 0.158 & 1.483 & 5.384 & 0.161 & 1.341 & 4.421 & 0.154 & $\textbf{4.494}$ & $\textbf{8.967}$ & 0.053 & 2.828 & 8.223 & 0.104 & 1.502 & 4.071 & 0.142 \\ 
				\specialrule{0em}{0pt}{6pt}
				\hline
				\specialrule{0em}{6pt}{0pt}
				Proposed-EDM & $\textbf{0.964}$ & $\textbf{4.318}$ & $\textbf{0.355}$ & $\textbf{1.040}$ & $\textbf{4.750}$ & $\textbf{0.352}$ & $\textbf{0.505}$ & $\textbf{3.365}$ & $\textbf{0.555}$ & 5.469 & 9.977 & 0.135 & $\textbf{2.371}$ & $\textbf{7.158}$ & $\textbf{0.266}$ & $\textbf{0.442}$ & $\textbf{2.400}$ & $\textbf{0.544}$ \\ 
				Proposed-CD & 0.982 & 4.342 & 0.344 & 1.058 & 4.802 & 0.341 & 0.521 & 3.449 & 0.547 & 5.337 & 9.954 & 0.133 & 2.412 & 7.209 & 0.260 & 0.454 & 2.487 & 0.542 \\ 
				\specialrule{0em}{6pt}{0pt}
				\hline
			\end{tabular}%
		}
		\vspace{-0.25cm}
	\end{table*}
	
	\begin{table}[ht]
		\captionsetup{justification=centering} 
		
		\caption{\\Quantitative Evaluation of the Proposed \\ Method  on Our Self-Made Dataset.}
		\label{tab:quantative_self}
		\centering
		\renewcommand{\arraystretch}{1.6}
		\setlength\tabcolsep{2pt}
		\resizebox{0.8\linewidth}{!}{%
			\fontsize{7}{6}\selectfont
			\begin{tabular}{l|ccc}
				\hline
				
				\specialrule{0em}{2pt}{0pt}
				\makebox[0.1\textwidth][l]{Methods} &  \makebox[0.07\textwidth]{CD \scalebox{1}{$\downarrow$}}  & \makebox[0.07\textwidth]{HD \scalebox{1}{$\downarrow$}}  & \makebox[0.07\textwidth]{F-Score \scalebox{1}{$\uparrow$}}  \\
				
				\specialrule{0em}{0pt}{2pt}
				\hline
				\specialrule{0em}{2pt}{0pt}
				
				OS-CFAR\cite{richards2005fundamentals} & 2.306 & 5.761 & 0.109 \\ 
				Proposed-EDM & $\textbf{0.669}$ & $\textbf{3.060}$ & $\textbf{0.325}$ \\ 
				Proposed-CD & 0.719 & 3.068 & 0.311 \\ 
				
				\specialrule{0em}{2pt}{0pt}
				
				\hline
			\end{tabular}%
		}
		\vspace{-0.5cm}
	\end{table}
	
	
	\begin{figure}
		\centering
		\begin{subfigure}{0.48\linewidth}
			\centering
			\includegraphics[width=1.05\linewidth]{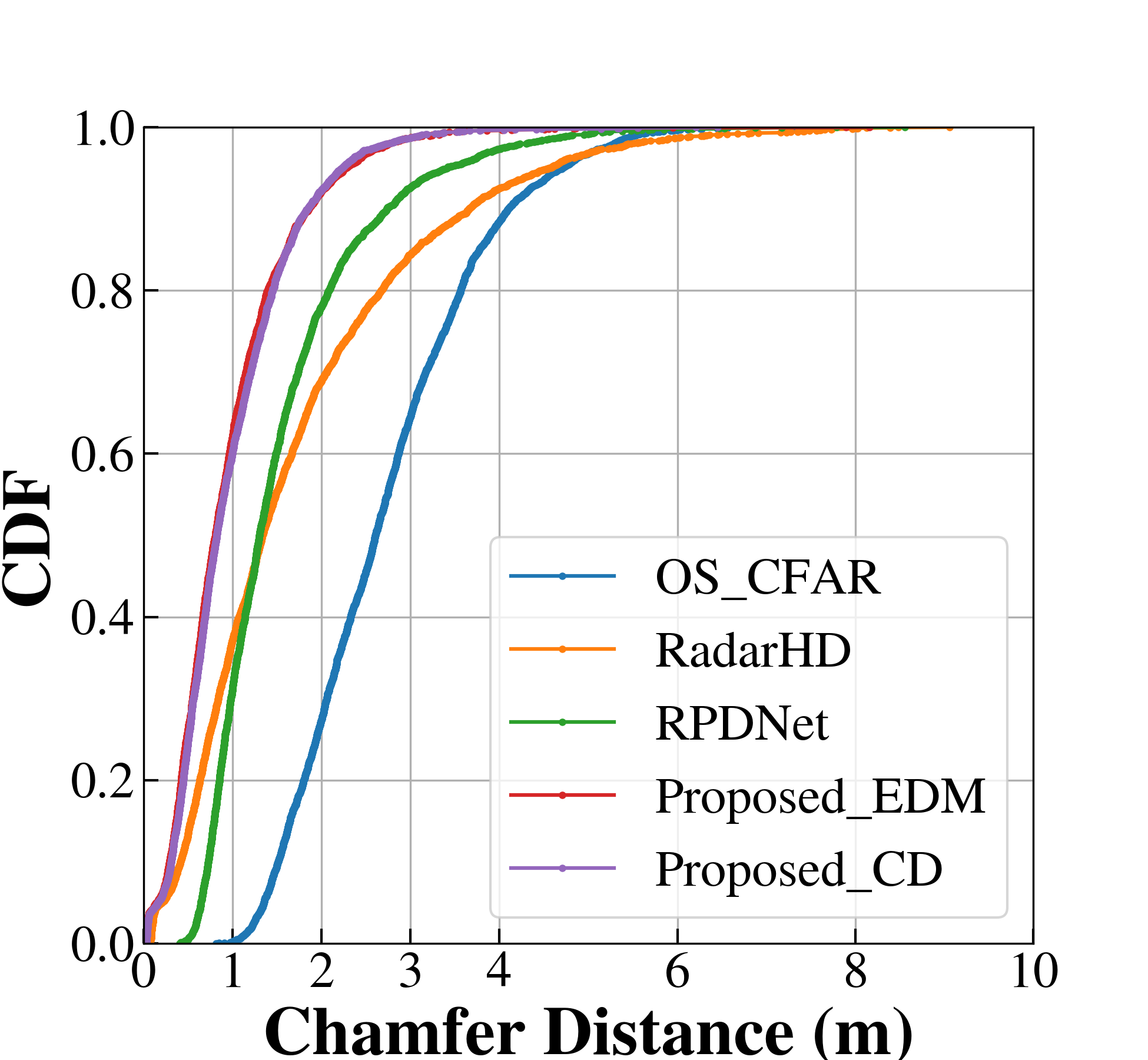}
			\caption{Arpg Lab}
			\label{pic:arpg}
		\end{subfigure}
		\centering
		\begin{subfigure}{0.48\linewidth}
			\centering
			\includegraphics[width=1.05\linewidth]{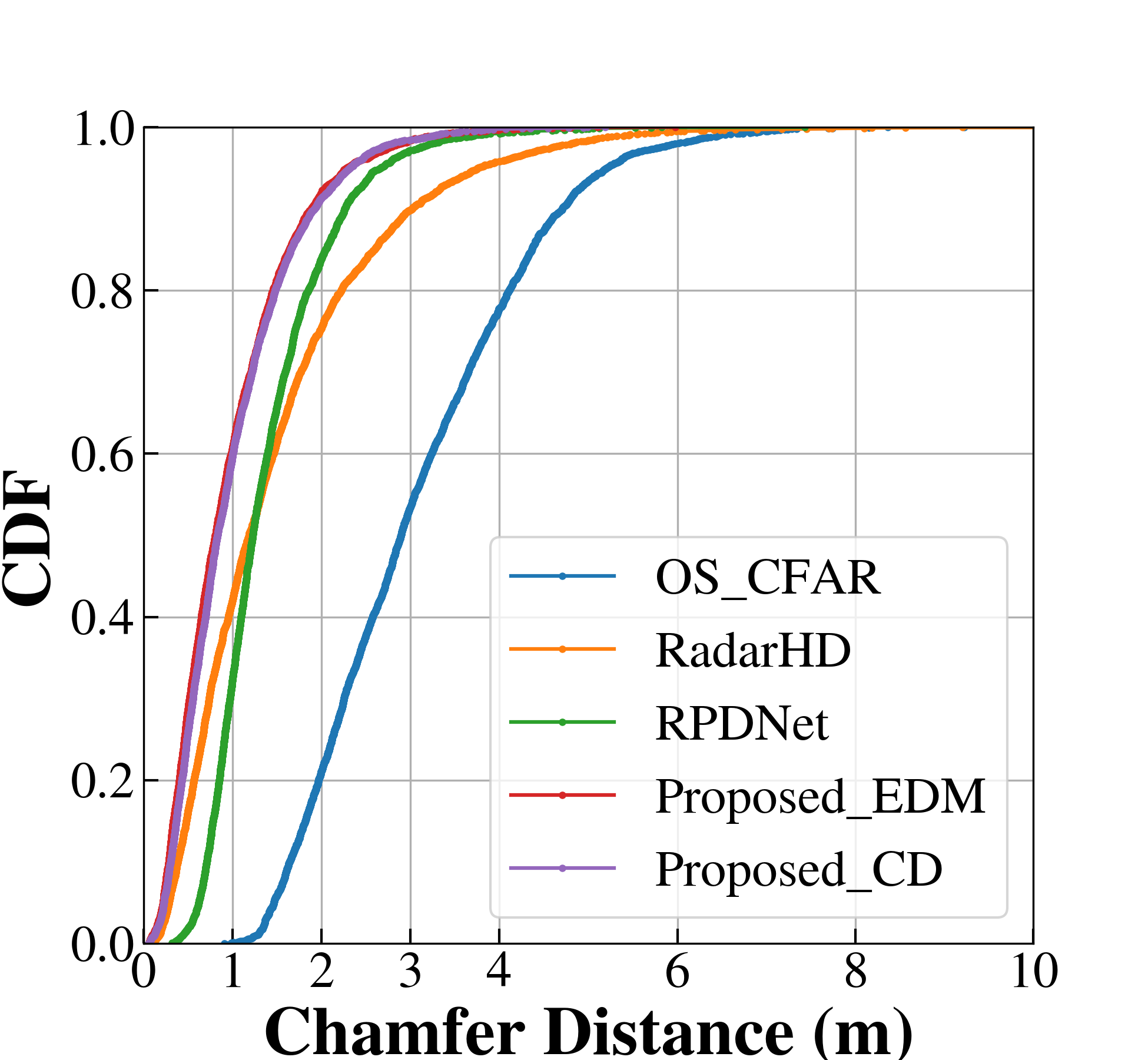}
			\caption{Ec Hallways}
			\label{pic:ec}
		\end{subfigure}
		\centering
		\begin{subfigure}{0.48\linewidth}
			\centering
			\includegraphics[width=1.05\linewidth]{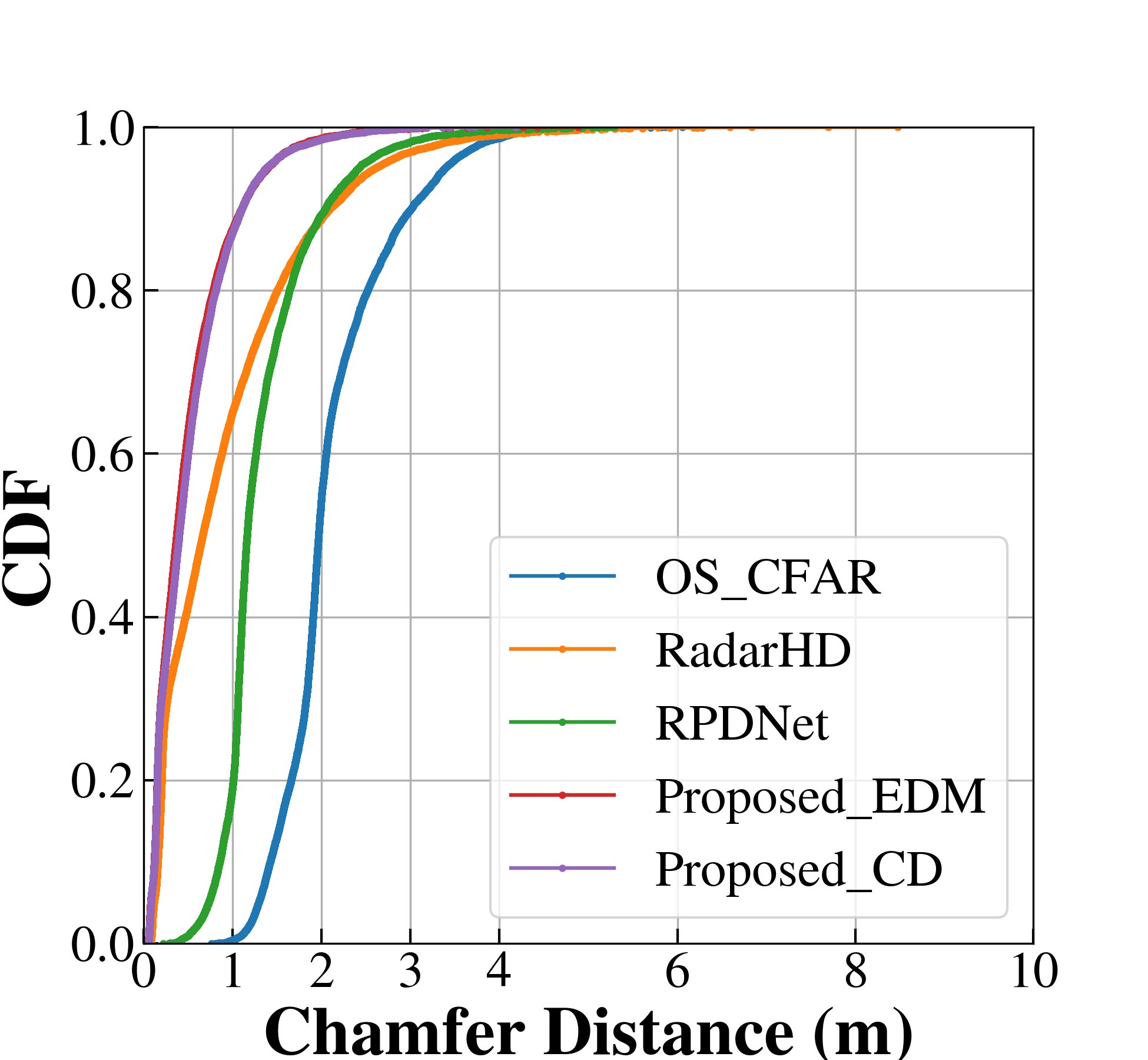}
			\caption{Aspen}
			\label{pic:aspen}
		\end{subfigure}
		\centering
		\begin{subfigure}{0.48\linewidth}
			\centering
			\includegraphics[width=1.05\linewidth]{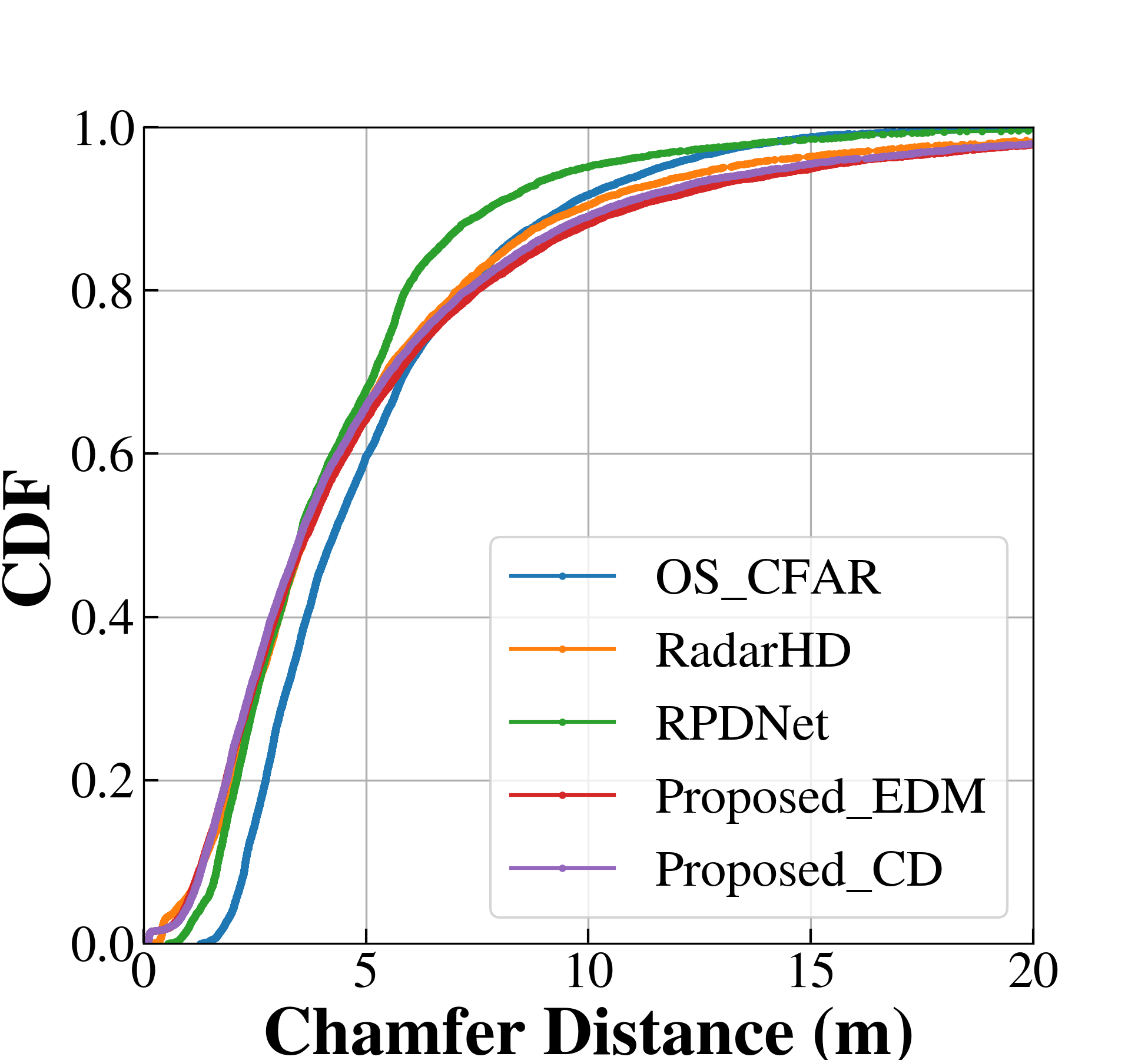}
			\caption{Longboard}
			\label{pic:longboard}
		\end{subfigure}
		\centering
		\begin{subfigure}{0.48\linewidth}
			\centering
			\includegraphics[width=1.05\linewidth]{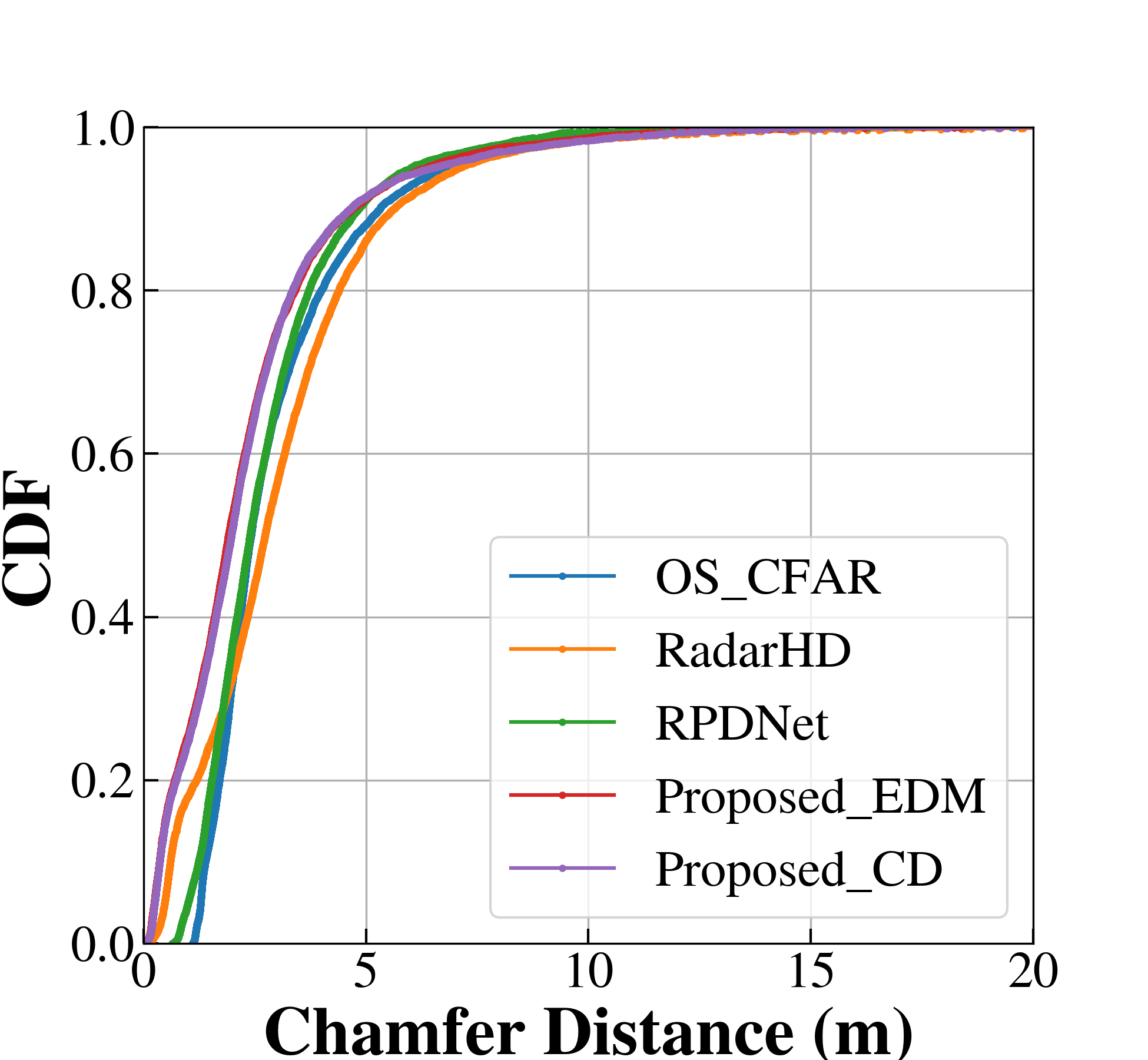}
			\caption{Outdoors}
			\label{pic:outdoors}
		\end{subfigure}
		\centering
		\begin{subfigure}{0.48\linewidth}
			\centering
			\includegraphics[width=1.05\linewidth]{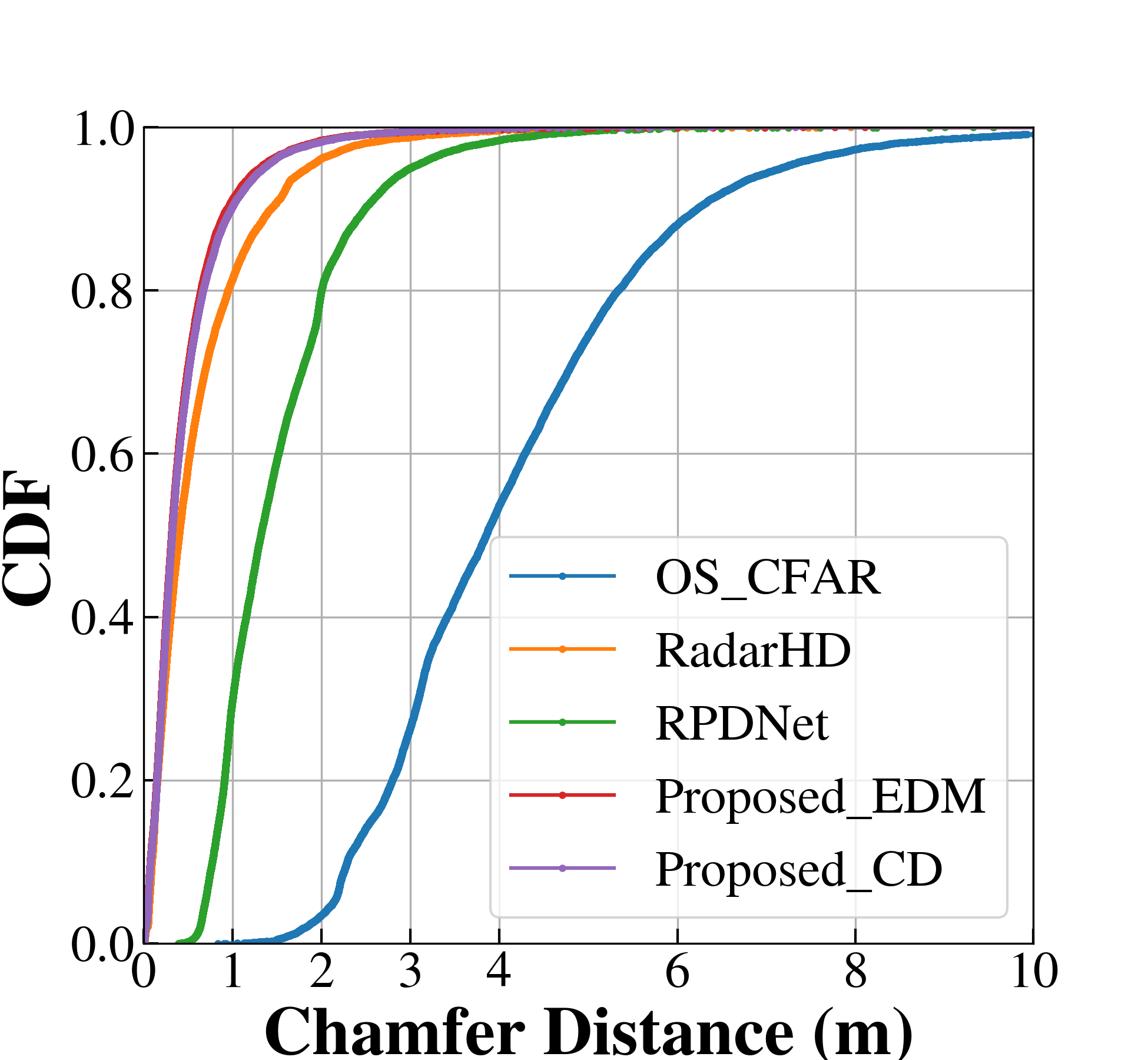}
			\caption{Edgar}
			\label{pic:edgar}
		\end{subfigure}
		\caption{CDF curves for the proposed method and the baseline methods on the ColoRadar dataset.}
		\label{pic:cdf_curves}
		\vspace{-0.7cm}
	\end{figure}

	Examples of ground truth LiDAR and radar point clouds generated by the proposed and baseline methods are shown in Fig.\ref{pic:qualitative_results}. The results reveal that the proposed method surpasses the baseline methods by generating denser and more accurate point clouds with fewer clutter points. Structural features of the scenes such as straight walls in the indoor scene, curved rock walls in the mine scene, and various obstacles in the Outdoor scene are also accurately predicted, which the baseline methods fail to do. Moreover, the results produced by our method in one-step generation are close to those obtained by 80 steps of iteration, indicating that consistency distillation achieves 80X acceleration without visually sacrificing the result quality.
	Among the baseline methods, the results of OS-CFAR and RPDNet contain a lot of clutter because they rely on traditional DOA estimation, which can hardly filter out noise in complex environments. Additionally, the angular resolution is much lower compared to the ground truth due to the absence of LiDAR supervision in the azimuth dimension. In contrast, RadarHD yields results with higher angular resolution and fewer clutter points because it adopts an end-to-end approach to learn the perception characteristics of LiDARs. However, due to its limited generation capacity, RadarHD cannot accurately preserve fine texture details in the environment, as the proposed method does. Furthermore, it fails to predict the environment structure in the outdoor scenes (both in Outdoor and Longboard). 
	In the Longboard dataset, the proposed and the baseline methods perform less effectively compared to other datasets  (as shown in Fig.\ref{pic:qualitative_results}). This is because the scenes in this dataset contain a significant amount of objects with low reflectivity, such as grass and leaves, which are hard to be detected by mmWave radars. In addition, these objects are highly unstructured and difficult to learn. As a result, the generated radar point clouds are sparser compared to LiDAR point clouds. We will further discuss this in the following quantitative comparison. More details are available in the supplementary video (or the extended one at https://www.youtube.com/watch?v=Q3S-9w3dGV4\&t=16s).
	
	\subsubsection{\textbf{Quantitative Comparisons}}
	We first introduce the evaluation metrics in the quantitative comparisons. We consider Chamfer distance (CD), Hausdorff Distance (HD), and F-Score\cite{tatarchenko2019single} in our case. We denote a ground truth point cloud set and a radar point cloud set as $\mathcal{P}_{GT}$ and $\mathcal{G}_{RD}$, respectively. 
	
	CD measures the dissimilarity between point cloud sets by calculating the sum of distances from each point to its nearest neighbor between $\mathcal{P}_{GT}$ and $\mathcal{G}_{RD}$. It is written as:
	
	\begin{equation}
		\frac{1}{\left|\mathcal{P}_{GT}\right|} \sum_{p \in \mathcal{P}_{GT}} \min _{g \in \mathcal{G}_{RD}}\|p-g\|+\frac{1}{\left|\mathcal{G}_{RD}\right|} \sum_{g \in\mathcal{G}_{RD}} \min _{p \in \mathcal{P}_{GT}}\|p-g\|.
	\end{equation}
	
	HD measures the longest distance from each point to its nearest neighbor between $\mathcal{P}_{GT}$ and $\mathcal{G}_{RD}$ and is defined as:

	\begin{equation}
		\max (D(\mathcal{P}_{GT}, \mathcal{G}_{RD}), D(\mathcal{G}_{RD}, \mathcal{P}_{GT})), 
	\end{equation}
	where
	\begin{equation}  D(A, B) = 
		\max_{a \in A}\{\min _{b \in B}\|a-b\| \}.
	\end{equation}
	
	F-Score evaluates the reconstruction quality and is defined as the harmonic mean between precision $P$ and recall $R$: 
	\begin{equation}
		\frac{2 P(\mathcal{P}_{GT}, \mathcal{G}_{RD}, d) \times 	R(\mathcal{P}_{GT}, \mathcal{G}_{RD}, d)}{P(\mathcal{P}_{GT}, \mathcal{G}_{RD}, d) + 	R(\mathcal{P}_{GT}, \mathcal{G}_{RD}, d)},
	\end{equation}
	where $d = 0.1 m$ is the distance threshold.
	
	The quantitative results are listed in Tab.\ref{tab:quantative_coloradar}. Fig.\ref{pic:cdf_curves} displays the cumulative distribution function (CDF) curves of chamfer distances. The closer the curve is to the vertical axis, the closer the generated point clouds are to the ground truth.
	It turns out that our approach (both Proposed-EDM and Proposed-CD) outperforms all baseline methods on quantitative metrics in every scenario except Longboard. These results convincingly demonstrate the superiority of our method in generating high-quality mmWave radar point clouds. As for the results in the Longboard scene, as described in section A, all methods perform poorly due to the presence of numerous objects that are challenging to be detected by mmWave radar. To address this issue, we will explore more reasonable learning mechanisms in the future to prevent the neural network from learning features that mmWave radar can hardly perceive. This will help to avoid misleading the optimization of the neural network.

	\vspace{-0.3cm} 
	\subsection{Real-world Experiments}
	In addition to conducting benchmark comparisons on the ColoRadar dataset, we also verify the proposed method using our self-made dataset. The main purpose is to test the generalization ability of the proposed method.
	\subsubsection{\textbf{Sensor Platform and Data Recording}}
	We first customize a hand-held sensor platform as illustrated in Fig.\ref{pic:sensor_platform} to record paired radar and LiDAR data. It consists of an Nvidia Jetson Orin NX embedded computer\footnote{https://www.nvidia.com/en-us/autonomous-machines/embedded-systems/jetson-orin/}, a Livox Mid-360 LiDAR\footnote{https://www.livoxtech.com/mid-360}, a TI-AWR1843BOOST single-chip mmWave radar\footnote{https://www.ti.com.cn/tool/en/AWR1843BOOST} with 3 TX and 4 RX antennas, and a TI-DCA1000EVM for raw radar data capture\footnote{https://www.ti.com.cn/tool/EN/DCA1000EVM}. We record a total of 3,152 frames of data in the hallway of Huzhou Institute of Zhejiang University. The radar and LiDAR data are synchronized through the recorded timestamps. Additionally, we utilize Fast-Lio\cite{xu2021fast}, running on the embedded computer, to record the poses corresponding to each frame of data for visualization purposes. Data preprocessing is the same as that in the benchmark comparisons as stated in Section \ref{sec:benchmark_dataset_preparation}.
	\subsubsection{\textbf{Generalization Ability Tests}}
	After data preparation, we use the proposed method's model trained on the ColoRadar dataset, without extra training, to directly infer the self-recorded radar data and compare it with the LiDAR ground truth.
	This test, which includes completely new scenes, a different LiDAR sensor, and the same mmWave radar with different parameters, presents a significant challenge to the generalization capability of the proposed method. For robustness, we concatenate the current and past ten frames of radar RAHs as the input to the network, and take the average of the predicted results as the output. The qualitative and quantitative results are presented in Fig.\ref{pic:realworld_test} and Tab.\ref{tab:quantative_self}. As can be seen, the proposed method manages to generate point clouds close to the ground truth and accurately predict the structural features. Even the quality of the results obtained by one-step generation is far superior to that of the traditional OS-CFAR algorithm. The above results strongly demonstrate the generalization ability and robustness of the proposed method and its practicability in autonomous navigation. More details are available in the supplementary video.
	\subsubsection{\textbf{System Efficiency}}
	We test the inference speed of the proposed method on a Jetson Orin NX with limited GPU computing capability. The average time consumption for one-step inference is 181.6 milliseconds, which implies a frequency of 5 Hz. This is sufficient to meet the real-time requirements for MAV autonomous navigation. Additionally, we can leverage techniques such as half-precision inference and TensorRT for further acceleration.

	\section{Conclusion}
	\label{sec:conclusion}
	This paper proposes a single-chip mmWave radar point cloud construction approach targeted for MAV autonomous navigation. Based on cross-modal supervision and generative learning utilizing aligned LiDAR point clouds and diffusion models,
	the proposed method can generate high-quality LiDAR-like point clouds from single-frame sparse and noisy radar data. Moreover, we incorporate the most recent diffusion model inference accelerating technology to achieve one-step generation and real-time inference on onboard computing platforms, thus addressing the slow inference issue of diffusion models. The proposed method excels the baseline methods in point cloud quality on the public ColoRadar dataset. Moreover, the generalization ability of the proposed method is verified against brand-new scenarios and different sensor configurations.
	
	In the future, we will first incorporate the elevation measurement information from mmWave radars to generate high-quality 3D point clouds. Then, we will deploy our method on an MAV to carry out autonomous navigation tasks in cluttered and visually degraded environments. 
	\section{Acknowlegments}
	\label{sec:acknowlegments}
	The authors would like to thank Yuwei Cheng for his help in raw radar data processing, dataset preparation, and the use of TI mmWave radar drivers.

	\bibliography{IROS2024_zrb}

\begin{thebibliography}{10}
\providecommand{\url}[1]{#1}
\csname url@samestyle\endcsname
\providecommand{\newblock}{\relax}
\providecommand{\bibinfo}[2]{#2}
\providecommand{\BIBentrySTDinterwordspacing}{\spaceskip=0pt\relax}
\providecommand{\BIBentryALTinterwordstretchfactor}{4}
\providecommand{\BIBentryALTinterwordspacing}{\spaceskip=\fontdimen2\font plus
\BIBentryALTinterwordstretchfactor\fontdimen3\font minus
  \fontdimen4\font\relax}
\providecommand{\BIBforeignlanguage}[2]{{%
\expandafter\ifx\csname l@#1\endcsname\relax
\typeout{** WARNING: IEEEtran.bst: No hyphenation pattern has been}%
\typeout{** loaded for the language `#1'. Using the pattern for}%
\typeout{** the default language instead.}%
\else
\language=\csname l@#1\endcsname
\fi
#2}}
\providecommand{\BIBdecl}{\relax}
\BIBdecl

\bibitem{richards2005fundamentals}
M.~A. Richards \emph{et~al.}, \emph{Fundamentals of radar signal
  processing}.\hskip 1em plus 0.5em minus 0.4em\relax Mcgraw-hill New York,
  2005, vol.~1.

\bibitem{park20213d}
Y.~S. Park, Y.-S. Shin, J.~Kim, and A.~Kim, ``3d ego-motion estimation using
  low-cost mmwave radars via radar velocity factor for pose-graph slam,''
  \emph{IEEE Robotics and Automation Letters}, vol.~6, no.~4, pp. 7691--7698,
  2021.

\bibitem{zhang20234dradarslam}
J.~Zhang, H.~Zhuge, Z.~Wu, G.~Peng, M.~Wen, Y.~Liu, and D.~Wang, ``4dradarslam:
  A 4d imaging radar slam system for large-scale environments based on pose
  graph optimization,'' in \emph{2023 IEEE International Conference on Robotics
  and Automation (ICRA)}.\hskip 1em plus 0.5em minus 0.4em\relax IEEE, 2023,
  pp. 8333--8340.

\bibitem{huang2023multi}
J.-T. Huang, R.~Xu, A.~Hinduja, and M.~Kaess, ``Multi-radar inertial odometry
  for 3d state estimation using mmwave imaging radar,'' \emph{arXiv preprint
  arXiv:2311.08608}, 2023.

\bibitem{chadwick2019distant}
S.~Chadwick, W.~Maddern, and P.~Newman, ``Distant vehicle detection using radar
  and vision,'' in \emph{2019 International Conference on Robotics and
  Automation (ICRA)}.\hskip 1em plus 0.5em minus 0.4em\relax IEEE, 2019, pp.
  8311--8317.

\bibitem{cheng2021robust}
Y.~Cheng, H.~Xu, and Y.~Liu, ``Robust small object detection on the water
  surface through fusion of camera and millimeter wave radar,'' in
  \emph{Proceedings of the IEEE/CVF International Conference on Computer
  Vision}, 2021, pp. 15\,263--15\,272.

\bibitem{wang2021rodnet}
Y.~Wang, Z.~Jiang, Y.~Li, J.-N. Hwang, G.~Xing, and H.~Liu, ``Rodnet: A
  real-time radar object detection network cross-supervised by camera-radar
  fused object 3d localization,'' \emph{IEEE Journal of Selected Topics in
  Signal Processing}, vol.~15, no.~4, pp. 954--967, 2021.

\bibitem{lu2020see}
C.~X. Lu, S.~Rosa, P.~Zhao, B.~Wang, C.~Chen, J.~A. Stankovic, N.~Trigoni, and
  A.~Markham, ``See through smoke: robust indoor mapping with low-cost mmwave
  radar,'' in \emph{Proceedings of the 18th International Conference on Mobile
  Systems, Applications, and Services}, 2020, pp. 14--27.

\bibitem{cheng2022novel}
Y.~Cheng, J.~Su, M.~Jiang, and Y.~Liu, ``A novel radar point cloud generation
  method for robot environment perception,'' \emph{IEEE Transactions on
  Robotics}, vol.~38, no.~6, pp. 3754--3773, 2022.

\bibitem{prabhakara2023high}
A.~Prabhakara, T.~Jin, A.~Das, G.~Bhatt, L.~Kumari, E.~Soltanaghai, J.~Bilmes,
  S.~Kumar, and A.~Rowe, ``High resolution point clouds from mmwave radar,'' in
  \emph{2023 IEEE International Conference on Robotics and Automation
  (ICRA)}.\hskip 1em plus 0.5em minus 0.4em\relax IEEE, 2023, pp. 4135--4142.

\bibitem{geng2023dream}
R.~Geng, Y.~Li, D.~Zhang, J.~Wu, Y.~Gao, Y.~Hu, and Y.~Chen, ``Dream-pcd: Deep
  reconstruction and enhancement of mmwave radar pointcloud,'' \emph{arXiv
  preprint arXiv:2309.15374}, 2023.

\bibitem{song2023consistency}
Y.~Song, P.~Dhariwal, M.~Chen, and I.~Sutskever, ``Consistency models,''
  \emph{arXiv preprint arXiv:2303.01469}, 2023.

\bibitem{goodfellow2020generative}
I.~Goodfellow, J.~Pouget-Abadie, M.~Mirza, B.~Xu, D.~Warde-Farley, S.~Ozair,
  A.~Courville, and Y.~Bengio, ``Generative adversarial networks,''
  \emph{Communications of the ACM}, vol.~63, no.~11, pp. 139--144, 2020.

\bibitem{kingma2013auto}
D.~P. Kingma and M.~Welling, ``Auto-encoding variational bayes,'' \emph{arXiv
  preprint arXiv:1312.6114}, 2013.

\bibitem{carvalho2023motion}
J.~Carvalho, A.~T. Le, M.~Baierl, D.~Koert, and J.~Peters, ``Motion planning
  diffusion: Learning and planning of robot motions with diffusion models,'' in
  \emph{2023 IEEE/RSJ International Conference on Intelligent Robots and
  Systems (IROS)}.\hskip 1em plus 0.5em minus 0.4em\relax IEEE, 2023, pp.
  1916--1923.

\bibitem{janner2022planning}
M.~Janner, Y.~Du, J.~B. Tenenbaum, and S.~Levine, ``Planning with diffusion for
  flexible behavior synthesis,'' \emph{arXiv preprint arXiv:2205.09991}, 2022.

\bibitem{schmidt1986multiple}
R.~Schmidt, ``Multiple emitter location and signal parameter estimation,''
  \emph{IEEE transactions on antennas and propagation}, vol.~34, no.~3, pp.
  276--280, 1986.

\bibitem{ding2023hidden}
F.~Ding, A.~Palffy, D.~M. Gavrila, and C.~X. Lu, ``Hidden gems: 4d radar scene
  flow learning using cross-modal supervision,'' in \emph{Proceedings of the
  IEEE/CVF Conference on Computer Vision and Pattern Recognition}, 2023, pp.
  9340--9349.

\bibitem{mopidevi2023rmap}
A.~N. Mopidevi, K.~Harlow, and C.~Heckman, ``Rmap: Millimeter-wave radar
  mapping through volumetric upsampling,'' \emph{arXiv preprint
  arXiv:2310.13188}, 2023.

\bibitem{sohl2015deep}
J.~Sohl-Dickstein, E.~Weiss, N.~Maheswaranathan, and S.~Ganguli, ``Deep
  unsupervised learning using nonequilibrium thermodynamics,'' in
  \emph{International conference on machine learning}.\hskip 1em plus 0.5em
  minus 0.4em\relax PMLR, 2015, pp. 2256--2265.

\bibitem{ho2020denoising}
J.~Ho, A.~Jain, and P.~Abbeel, ``Denoising diffusion probabilistic models,''
  \emph{Advances in neural information processing systems}, vol.~33, pp.
  6840--6851, 2020.

\bibitem{song2019generative}
Y.~Song and S.~Ermon, ``Generative modeling by estimating gradients of the data
  distribution,'' \emph{Advances in neural information processing systems},
  vol.~32, 2019.

\bibitem{song2020score}
Y.~Song, J.~Sohl-Dickstein, D.~P. Kingma, A.~Kumar, S.~Ermon, and B.~Poole,
  ``Score-based generative modeling through stochastic differential
  equations,'' \emph{arXiv preprint arXiv:2011.13456}, 2020.

\bibitem{kawar2022denoising}
B.~Kawar, M.~Elad, S.~Ermon, and J.~Song, ``Denoising diffusion restoration
  models,'' in \emph{Advances in Neural Information Processing Systems}, 2022.

\bibitem{karras2022elucidating}
T.~Karras, M.~Aittala, T.~Aila, and S.~Laine, ``Elucidating the design space of
  diffusion-based generative models,'' \emph{Advances in Neural Information
  Processing Systems}, vol.~35, pp. 26\,565--26\,577, 2022.

\bibitem{lin2023diffusion}
S.~Lin and X.~Yang, ``Diffusion model with perceptual loss,'' \emph{arXiv
  preprint arXiv:2401.00110}, 2023.

\bibitem{zhang2018unreasonable}
R.~Zhang, P.~Isola, A.~A. Efros, E.~Shechtman, and O.~Wang, ``The unreasonable
  effectiveness of deep features as a perceptual metric,'' in \emph{Proceedings
  of the IEEE conference on computer vision and pattern recognition}, 2018, pp.
  586--595.

\bibitem{ronneberger2015u}
O.~Ronneberger, P.~Fischer, and T.~Brox, ``U-net: Convolutional networks for
  biomedical image segmentation,'' in \emph{Medical Image Computing and
  Computer-Assisted Intervention--MICCAI 2015: 18th International Conference,
  Munich, Germany, October 5-9, 2015, Proceedings, Part III 18}.\hskip 1em plus
  0.5em minus 0.4em\relax Springer, 2015, pp. 234--241.

\bibitem{nichol2021improved}
A.~Q. Nichol and P.~Dhariwal, ``Improved denoising diffusion probabilistic
  models,'' in \emph{International Conference on Machine Learning}.\hskip 1em
  plus 0.5em minus 0.4em\relax PMLR, 2021, pp. 8162--8171.

\bibitem{kramer2022coloradar}
A.~Kramer, K.~Harlow, C.~Williams, and C.~Heckman, ``Coloradar: The direct 3d
  millimeter wave radar dataset,'' \emph{The International Journal of Robotics
  Research}, vol.~41, no.~4, pp. 351--360, 2022.

\bibitem{lee2022patchwork++}
S.~Lee, H.~Lim, and H.~Myung, ``Patchwork++: Fast and robust ground
  segmentation solving partial under-segmentation using 3d point cloud,'' in
  \emph{2022 IEEE/RSJ International Conference on Intelligent Robots and
  Systems (IROS)}.\hskip 1em plus 0.5em minus 0.4em\relax IEEE, 2022, pp.
  13\,276--13\,283.

\bibitem{tatarchenko2019single}
M.~Tatarchenko, S.~R. Richter, R.~Ranftl, Z.~Li, V.~Koltun, and T.~Brox, ``What
  do single-view 3d reconstruction networks learn?'' in \emph{Proceedings of
  the IEEE/CVF conference on computer vision and pattern recognition}, 2019,
  pp. 3405--3414.

\bibitem{xu2021fast}
W.~Xu and F.~Zhang, ``Fast-lio: A fast, robust lidar-inertial odometry package
  by tightly-coupled iterated kalman filter,'' \emph{IEEE Robotics and
  Automation Letters}, vol.~6, no.~2, pp. 3317--3324, 2021.

\end{thebibliography}
\end{document}